\documentclass[10pt, a4paper]{article}

\usepackage[final]{lrec2026} 

\usepackage[english]{babel}
\babelprovide[import]{chinese}
\babelfont[chinese]{rm}{NotoSansCJK-Regular.ttc} 
\babelprovide[import]{korean}
\babelfont[korean]{rm}{NotoSansCJK-Regular.ttc} 
\babelprovide[import]{hindi}
\babelfont[hindi]{rm}{NotoSansDevanagari-Regular.ttf} 
\babelprovide[import]{farsi}
\babelfont[farsi]{rm}{NotoSansArabic-Regular.ttf} 

\usepackage{booktabs}
\usepackage[shortlabels]{enumitem}
\usepackage{todonotes}

\usepackage{pifont}
\usepackage[dvipsnames]{xcolor}
\newcommand{\badmark}{\textcolor{Red}{\ding{54}}}  
\newcommand{\goodmark}{\textcolor{Green}{\ding{52}}} 

\newcommand{\benchmark}{Tuguesice-PT}

\title{Progressing beyond Art Masterpieces or Touristic Clichés:\\ how to assess your LLMs for cultural alignment?}

\name{%
António Branco,$^\dag$ João Silva,$^\dag$ Nuno Marques,$^\dag$ Luis Gomes,$^\dag$ Ricardo Campos,$^\ddag$ \\
{\bf\large Raquel Sequeira,$^\ddag$ Sara Nerea,$^\ddag$ Rodrigo Silva,$^\ddag$ Miguel Marques,$^\ddag$} \\
{\bf\large  Rodrigo Duarte,$^\ddag$ Artur Putyato,$^\ddag$ Diogo Folques,$^\ddag$ Tiago Valente$^\ddag$}} 

\address{$^\dag$University of Lisbon, $^\ddag$University of Beira Interior \\
         antonio.branco@di.fc.ul.pt\\}

\abstract{
Although the cultural (mis)alignment of Large Language Models (LLMs) has attracted increasing attention---often framed in terms of cultural bias---until recently there has been limited work on the design and development of datasets for cultural assessment. Here, we review existing approaches to such datasets and identify their main limitations. To address these issues, we propose design guidelines for annotators and report on the construction of a dataset built according to these principles. We further present a series of contrastive experiments conducted with this dataset. The results demonstrate that our design yields test sets with greater discriminative power, effectively  distinguishing between models specialized for a given culture and those that are not, \textit{ceteris paribus}.
 \\ \newline \Keywords{LLMs, cultural alignment, test datasets, language resources evaluation} }

\begin{document}

\maketitleabstract

\section{Introduction}

Since the advent of Large Language Models (LLMs), the question of how to undertake their assessment has become a central topic of research. 
Initially such evaluation efforts were supported mostly by the so-called instruct datasets, consisting of pairs of input questions and of the respective gold answers, thus focusing on the semantic aptitude of the models, that is, in their aptitude to basically handle knowledge about the world \citep{ni2025surveylargelanguagemodel}.
Soon, further efforts were directed to other aptitudes as well \citeplanguageresource{mousi2024aradicebenchmarksdialectalcultural}, such as, among others, their aptitude in terms of civility, to have relations with humans or other agents within the limits of consensually established social constraints on courtesy and proper interaction~\citeplanguageresource{Wang:2023:donotanswer}.

Concomitantly, issues related to sovereignty became also a major topic of debate and research. 
As these have been articulated in a wide range of sectors, from governments to scientific researchers, they have been concerned with what has been termed ``cultural'', ``linguistic'', or, even more broadly, ``digital'' sovereignty \citep{radu2021, Mugge02082024}.

At the intersection of these two important lines of debate and inquiry, one finds the discussion on how to properly assess LLMs for cultural aptitude \citep{liu2024}, for their ability to know, acknowledge, handle, deliver, respect and/or support what any given group assumes it is or belongs to its own culture, which helps to individuate such group and socially bind their members among themselves---be it that it envisages itself as a nation, a people, a town, a group of fans, etc. 
Like for the assessment of other aptitudes of LLMs, a central role is played by the research on how to design and develop test datasets to undertake such assessment \citep{Khan2025}. 
That is the topic of this paper.

A first goal here is to present a critical analysis of the major examples or approaches in the literature to the design of test datasets for the purpose of evaluating cultural alignment. 
Our tenet is that they suffer from a number of limitations. 
Focusing mostly on historical events or literate culture highlights, that many times belong also to universal or global culture, or echoing stereotypical \textit{clichés} from points of view that are external to the social group at stake, among others, are just a couple of such limitations. 
Accordingly, the leverage of such datasets to ultimately fulfill its intended role---of discriminating between models that may happen to be more or less aligned with the distinctiveness of the culture of a group---vanishes given the widespread presence in the web of such facets, from where the training data for those models have been extensively scraped.

Taking that analysis into account, another major goal of this paper is to propose a design approach for these datasets that avoids such limitations. 
This approach is materialized in a set of development guidelines that we present and justify. 

Their appropriateness to provide discriminative leverage is empirically assessed by experimenting with the development of a dataset under such guidelines, concerning a given culture, and with its subsequent usage for evaluating the respective cultural alignment of a range of LLMs---large and small, with and without having been subject to dedicated finetuning for that culture.

We discuss previous approaches to the design of datasets for cultural alignment in Section~\ref{Sect:background}. 
Taking into account the lessons learned from that analysis, in Section~\ref{Sect:guidelines} we propose and discuss design guidelines for annotators, and in the following Section~\ref{Sect:dataset} the development of a dataset according to these guidelines is presented. 
This dataset was used to evaluate the cultural alignment of a range of LLMs and the results and implications of this experiment are reported in Section~\ref{Sect:assessment}. 
In the final Section~\ref{Sect:conclusions}, conclusions are presented.

\section{Background}
\label{Sect:background}

In this section we proceed with an overview of the literature related to the design of datasets aimed at assessing cultural alignment. Though the issues raised around this topic have received considerable highlight in the media and public discussion since the advent of ChatGPT, under the consideration of so-called cultural biases of models, somewhat surprisingly until a few years ago there were not many publications on cultural assessment datasets in the literature.

\subsection*{Translating and getting generic clichés}

The Benchmark for LLMs on Everyday Knowledge in Diverse Cultures and Languages (BLEnD), described in \citeplanguageresource{Myung:2024:blend}, is a benchmark for assessing cultural alignment that seems to have gathered considerable traction given the attempts to translate it into other languages. This is a dataset with over 52k question-answer pairs, with some questions coupled with short-answers and some others with multiple-choice answers. It includes multiple subsets, with the respective entries written in 13 languages, that are meant to address multiple cultures, from 16 countries.

The building process of BLEnD proceeds as follows:
Native speakers from different countries are asked to write a few culturally-relevant question-answer pairs.
In order to extend these pairs to other cultures,
each question is translated into the target language and the result is edited such that the country mentioned in that question is now the country of interest; this transposed question is coupled with its answer, newly formed for this purpose.

For instance, a Spanish annotator could have contributed the question ``What is the most popular indoor sport in Spain?''\footnote{¿Cuál es el deporte de interior más popular en España?}, together with the respective answer.
This question can then serve as a sort of template to generate similar questions for other cultures, by translating it into other languages and replacing ``España'' by another country name.
These newly-generated questions are then answered by the annotators that are native to the culture specified in the question, and this new question-answer is entered into the benchmark.

When empirically experimenting with actual LLMs on these datasets, these authors report that, as expected, while the LLMs tested tend to show good results for languages that are highly represented in the training data, performance drops sharply when LLMs answer questions pertaining to the cultures of low-resource languages.

We find a number of questionable assumptions and related drawbacks underlying the design of BLEnD. 
First, its questions tend to be such that they should receive as answers sentences that are generics---in the sense of ``generics´´ for grammatical analysis \citep{Pelletier1997-PELGAD}. 
Accordingly, as it is well known from the linguistics of generic sentences, about the ``soft´´ and ``tricky´´ generalizations they may express, these may have exceptions and may help to bias one's understanding: about the same subject, different persons tend to perceive different ``soft'' generalizations, and thus find that different answers may be better suited for a question, and do not consensually agree on a single answer. 
For instance, the question Ca-sp-19 is ``What do young people from Spain usually drink at the night club?''.\footnote{¿Qué suelen beber los jóvenes de España en la discoteca?}

Above all, as they do not require factual objective answers, these kind of questions and their answers tend to slip into accommodating stereotyped views on a given culture, e.g. ``Italians are good skiers''. 
Some answers tend to approximate what may appear as picturesque for external (from other cultures) observers of that culture, under the form of usual touristic \textit{clichés}: ``What do people do to celebrate New Year's Day in Greece?''.\footnote{Entry Al-en-34: Τι κάνουν οι άνθρωποι για να γιορτάσουν την Πρωτοχρονιά στην Ελλάδα?}

Second, translating into another language and transposing the question-answer pair to another national setting eventually leads to distortions. Some resulting questions are not really suitable, like asking what Chinese people usually eat for Thanksgiving, a holiday not celebrated in China.\footnote{Entry Al-en-32: \foreignlanguage{chinese}{在中国，感恩节的主菜是什么？}}

Third, when addressing a chatbot, people do not prefix their questions with the equivalent of the expressions ``in the US'', ``in Spain'', etc., to refer to which country or culture they belong to. 
And it is crucial in testing the cultural alignment of an LLM to see what it will answer without the help of that linguistic crutch, and what can thus be properly made explicit about its (dis)alignment with respect to a given culture. 
For instance, if to a question like ``What is the nuclear family?'', an LLM under assessment would eventually answer ``A family unit consisting of two parents and their children living together in one household.'',\footnote{duckduckgo.com AI assistant on Oct.~19, 2025} this is a sign that it is likely misaligned from the Chinese culture.

Given the above considerations, and in line with \citep{ramesh-etal-2023-fairness}, it should have had the immediate effect of raising one's eyebrows the fact that a data set aimed at assessing the aptitude of LLMs for different individual cultures contains mostly the same questions for every culture only that they were translated into different languages and prefixed with different country names.

\subsection*{Hunting for encyclopedic knowledge and ending up with low discriminative power}

On a par with the additions to BLEnD, in the past few months a flurry of datasets have been published that are designed under a common approach, namely of being made of entries concerned with some sort of factual encyclopedic knowledge about a given country, concerning for instance its geography, history, literature, etc. 
While they do not thus incur into the above ``touristic \textit{cliché}'' pitfall, they nevertheless suffer from some other drawbacks.

First, the questions tend not to focus on what people see as being part of their culture, but rather on encyclopedic knowledge, of all sorts, about their country. 
For example, it is not because Portuguese people mainly live in buildings with an average of 3 to 7 floors that they consider this accidental circumstance to be part of their culture.

Second, while the same pool of questions are not translated into different languages to form datasets aimed at addressing different cultures---thus avoiding similar pitfall in BLEnD---, they do tend to follow the same taxonomy to induce question writing for different cultures (e.g.~questions about history, food, holidays, etc.). 
This reinforces the approximation to a given culture from a uniform point of view that is external to it.

Third, being encyclopedic, and thus public, in its nature, the type of knowledge that the question-answer pairs seek to capture tends to be available on the web in some form. 
As a consequence, the resulting datasets tend to bear a meager discriminative power to actually tell apart LLMs 
in terms of their alignment to a given culture.
And this affects either well or less resourced languages and cultures, as the same amount of encyclopedic knowledge about a given culture is available to every model, as they are all trained on data scrapped from the web. 
Even when the questions may be about cultural aspects---rather than other type of knowledge about the country---, given their scope tend to be the so-called ``high culture'' (e.g.~art masterpieces, novelists, Nobel laureates, etc.), the same effect is observed, as this type of data tends also to be fairly documented on the web.

A minimal version of this approach can be found in \citeplanguageresource{mousi2024aradicebenchmarksdialectalcultural}.
Besides asking for the indication of 5 URLs substantiating the gold answer to each question, the only other guideline for annotators is: ``A question [in the respective language/dialect] that asks for information related to a specific country.'' (p.30). 
With the resulting dataset of just 30 questions on Persian public holidays, food, geography, and religion, the best model tested, with only 13B parameters, already scores well over 70\%.

SaudiCulture \citeplanguageresource{ayash2025saudiculturebenchmarkevaluatinglarge}, in turn, contains 441 questions under an explicit taxonomy, covering food, clothing, language, entertainment, celebrations, dating, crafts, and architecture. 
The best model experimented with is GPT4, scoring already well over 50\% in all subsets except one.

With 17,411 entries, spread over 22 subsets relating to 22 countries, on science, food, sports, politics, religion, history, travel, flora \& environment, geography, celebrations, language, and proverbs, the dataset reported in \citeplanguageresource{alwajih2025palmculturallyinclusivelinguistically} pushes further this type of approach. 
1,916 entries were sampled for empirical assessment, with the best performing model, Claude 3.5 Sonnet, scoring already well over 60\%, without having been fine-tuned for this task.

In \citeplanguageresource{zhang2025culturesynthhierarchicaltaxonomyguidedretrievalaugmented} the question taxonomy is deepened, covering 12 primary domains (Social Sciences, Philosophy and Psychology, Religion and Theology, Political Science, Law, Education, Language, Literature, Medicine, Applied Sciences and Technology, Arts, Recreation, Sports, and Entertainment), each articulated into 130 further secondary topics. 
A Retrieval-Augmented Generation (RAG)-based methodology leveraging factual knowledge was used to synthesize culturally relevant question-answer pairs in 7 languages. 
CultureSynth-7, the dataset that was thus automatically created, was used to evaluated 14 LLMs. 
As expected, its discriminative strength is low, with the best model achieving already more than 75\% accuracy, and the others following closely to it.

\citetlanguageresource{zhang2025culturescopedimensionallensprobing} go along the same approach, but now with a taxonomy with 140 categories, for two languages (Chinese and Spanish). 
The performance of most of the 11 LLMs put to test also reveal that this is an almost exhausted dataset, already at its inception, with between 70\% to 90\% scores.

Seeking to address the cultures in 15 languages of India, \citetlanguageresource{maji2025drishtikonmultimodalmultilingualbenchmark} introduce the novelty of including images and each entry in the dataset being made of triples question-answer-image. 
Again, for powerful LLMs, the resulting dataset shows residual discriminative strength with these models reaching already between 70\% to 80\% accuracy.

Interestingly, besides the LLMs performance scores, \citetlanguageresource{moosavi-monazzah-etal-2025-percul} also report on layperson's performance on the cultural dataset they developed under this approach. 
At the inception of the developed dataset, only a 11.3 percentage points gap exists from the best model to the respective human upper bound.

Stepping aside from the capture of encyclopedic knowledge and high culture, and tackling 7 themes ranging from food preferences to greeting etiquette, \citetlanguageresource{chiu-etal-2025-culturalbench} developed CulturalBench, with 1,696 human written questions, covering 45 global regions including underrepresented ones like Bangladesh, Zimbabwe and Peru. 
And compared to human performance upper bound (92.4\% accuracy), this dataset is challenging even for the best-performing frontier LLMs, ranging from 28.7\% to 61.5\% in accuracy. 
Nevertheless, the discriminative power of such kind of dataset is not as strong as it could be given that, as these authors report, they ``find that LLMs often struggle with tricky questions that have multiple correct answers (e.g.~``What utensils do the Chinese usually use?'') (p.25663). 
That is, they end up affected, in turn, by the pitfall discussed in the previous subsection, related to the usage of generic sentences and their expression of potentially stereotyped knowledge.

\subsection*{Trivia in forums or the news}

Possibly anticipating the limitation of hunting for well-established encyclopedic knowledge about a given country --- and the circumstance that this is somewhat long lasting and may be available online ---, \citetlanguageresource{arora-etal-2025-calmqa} turn to more recent and transient pieces of knowledge and build CaLMQA, with over 50k question-answer pairs, by inducing human written questions for 23 languages from web forums and the conversation found therein.

Some examples of questions (p.11773): 
``Which high school has a higher competition rate, Gyeonggi Foreign Language High School or Suwon Foreign Language High School?" in Korean,\footnote{\foreignlanguage{korean}{경기외고 수원외고 어느 고등학교가 경쟁률이 높은가요?}}
``Why did the Indian rocket PSLV-C39 fail to carry the satellite?" in Hindi,\footnote{\foreignlanguage{hindi}{भारतीय रॉकेट} PSLV-C39 \foreignlanguage{hindi}{भसेटेलाइट को ले जाने म फेल ों आ है?}}
or ``Where is Sleeping beauty mountain and how does it impact the tourism landscape?'' in Balochi,\footnote{\foreignlanguage{farsi}{ت؟ﯾﺑ دازﻧر اﺛوں اُ ہﻣﺎﻧظرﻧﻣ ردگ وّرﺗ و تﻧاِ ﺎﺟُک یٹوﯾﺑ گﻧپﯾﻠﺳ}}
among others.

To a large extent, datasets with these type of sources face the same drawbacks discussed in the sections above.
The questions tend not to focus on what people see as being part of their culture and the knowledge that the question-answer pairs capture tends to be available on the web. As a consequence the resulting datasets do not have an as strong discriminative power as they could. 
When used to assess the alignment of LLMs, these datasets show performance scores already close to or over 50\%, despite around half of the questions concern languages that are very low-resourced (p.11786).\footnote{Afar, Faroese, Fijian, Hiligaynon, Kirundi, etc.}

\subsection*{Narrowing into politeness procedures}

TaarofBench~\citeplanguageresource{Sadr:2025:politelyinsist} is a benchmark of role-play scenarios that assesses whether models are able to follow the Persian politeness rituals of taarof. The authors find that, as expected, the tested LLMs fare worse than humans in recognizing situations where taarof is appropriate, likely because in their training they have been mostly exposed to data from other languages, and other expectations with respect to human behavior.

While TaarofBench is focused on a very particular type of social interaction of a given culture, when it comes to cultural alignment, one seeks to assess with our datasets a much wider range of culture-specific background from our LLMs rather than just politeness conventions.

\subsection*{Narrowing into lists of proverbs}

Other articles also put forward proposals that represent a similar kind of narrowing into a very specific part of a given culture, which ultimately is too specific for the typical purpose of supporting a sensible assessing of cultural alignment.

The papers \citeplanguageresource{Almeida:2025:broverbs} and \citeplanguageresource{Gromenko:2025:russiancatchphrases} are two such cases, which happen to narrow down the dataset into just the list of proverbs in the languages they address, Portuguese and Russian respectively.

\subsection*{Narrowing into named entities}

BertaQA~\citeplanguageresource{Etxaniz:2024:bertaqa} is a question-answering benchmark, with 4.7k multiple-choice questions, parallel between English and Basque, split into two parts, one with questions that are specific to the Basque culture and another with questions that are deemed to be of global relevance. 
Experiments reported indicate that, as expected, the existing general-purpose LLMs perform better on the questions on global topics than on the Basque-specific questions.

We find two main weakness in the approach adopted for BertaQA. 
On the one hand, the questions are multiple-choice, with the correct answer among the choices. 
A fully-generative approach to generate the answer given only the question is clearly a setup that comes closer to providing a real challenge that may bring to light the level of cultural alignment of the model, rather than just selecting one out of the available options.
On the other hand, and more importantly, the Basque-specific questions tend to include named entities, such as the names of people or locations, that are specific to that culture. 
This way the question already provides important clues that lean the answer to align with the culture to which the named entities belong and likely inflate the performance of the model, hindering its fair scoring.

In this context, it is worth noting the proposal of \citetlanguageresource{zhao2025makievalmultilingualautomaticwikidatabased}, also relying on named entities. 
Their dataset is obtained by generating text and then counting the named entities belonging to a given language/culture occurring therein---rather than via the contrast against gold question-answer pairs, as in all other approaches discussed above.
Different runs of the generation of texts are done with a few different prompts, in different languages, with respect to different countries, and by a number of different LLMs, thus generating, as expected, different numbers of named entities that occur in the snippets generated. 
Interestingly, the authors acknowledge that ``how these differences should be interpreted and applied in practice remains an open question that merits further discussion'' (p.9).

\subsection*{Going astray with statistics from surveys}

A radical departure from collecting gold question-answer pairs to form a dataset to test LLMs is defended by \citet{alkhamissi-etal-2024-investigating}:
``We quantify cultural alignment by simulating sociological surveys, comparing model responses to those of actual survey participants as references'' (p.12404).

They selected 30 questions from the WVS-7 questionnaire \citeplanguageresource{Haerpfer:2020:wvs7} that aims at gathering responses to an array of multiple-choice questions on matters of ``social, cultural, material, governmental, ethical, and economic importance'' designed to include indicators towards several United Nations Sustainable Development Goals. 
These questions were run on a sample of human subjects and the answers grouped according to subject's gender, education level, social class, and age range. 
LLMs were prompted with the same answers and their answers were compared to the answers and distribution of scores from the subjects.

A similar radical departure is defended also by \citet{sukiennik2025evaluationculturalvaluealignment}. 
But here the key assumption is that a given culture is characterized by the results from surveys organized along the six dimensions of the questionnaire in \citep{Hofstede01121980}, namely Power Distance, Individualism vs.\ Collectivism, Masculinity vs.\ Femininity, Uncertainty Avoidance, Long-Term Orientation vs.\ Short-Term Orientation, and Indulgence vs.\ Restraint. 

Whatever may be the empirical scoring found, this type of approach goes astray with respect to our goal of assessing cultural alignment. What is considered to be part of the culture acknowledged by a group can be, and it is in most of the cases, orthogonal with respect to the different social class or to the education levels of the members of that culture. Just to provide an illustration, in Portugal fado music is appreciated irrespectively of the education levels of fado lovers. These drawbacks are extensively examined in \cite{Khan2025}.

Furthermore, this type of approach implies that one model can only handle at its best one culture, which admittedly does not have to be the case.

\section{Development guidelines}
\label{Sect:guidelines}

With the exception of the paper mentioned above that indicates annotation guidelines \citeplanguageresource{mousi2024aradicebenchmarksdialectalcultural}, a common trait of the related work reviewed is that no guidelines for annotators were presented. 
Decades of language resources development have informed us, however, that well-defined guidelines are essential to the aim, consistency and reliability of datasets. 

Consequently, we designed guidelines that support the development of datasets for assessing cultural alignment and that take into account the lessons learned with the literature review in Section \ref{Sect:background}. The guidelines are written on the assumption that the goal is for the dataset to have entries created manually by human annotators, following these guidelines, who are native speakers of the language in which the entries are written and who were raised and live in a particular culture, and are therefore fully immersed in and familiar with it.

\subsection{Linguistic constraints}
\label{Sect:linguistic}

A first set of guidelines include linguistic constraints on the questions and answers that are allowed to integrate the dataset, seeking to avoid the limitations discussed, including those related to questions that ask for generic sentences as answers, and eventually induce stereotypes. 
Given the dataset is aimed at supporting an automatic evaluation process, it is also important that the questions facilitate it by having only short, non-list answers.

To help clarify the purpose of the guidelines, each instruction is associated with a pair of sentences that are examples that comply and fail to comply with that instruction.\footnote{Such examples are provided to the annotators in the target language. They are present here in English to enhance readability.}
Many of the bad examples presented below are taken from entries that were actually proposed by annotators (and rejected upon adjudication).

\begin{enumerate}[L1.]\small

\item Linguistic complexity of the question: The question shouldn't be too long, or contain subordinate clauses, rare words, etc.

    \badmark\ \textit{Taking into account the evolution of public debt and considering the trajectory of inflation, who was the most successful finance minister?}\\    
    \goodmark\ \textit{Who was the finance minister during the COVID pandemic?}
    
\item Linguistic complexity of the answer: The answer should be a single item, and not a list.
    
    \badmark\ \textit{What are the main tourist attractions in our country?}\\    
    \goodmark\ \textit{What is the capital of stone soup?}
    
    
\item Factuality: The question should be factual.
    
    \badmark\ \textit{Who is right in the debate over the legalization of abortion?}\\    
    \goodmark\ \textit{In what year was the most recent bridge over the river inaugurated in the capital?}
    
\item Single answer: The question has to be unambiguous and have only one possible answer.
    
    \badmark\ \textit{What is the main form of public transportation in large cities?}\\    
    \goodmark\ \textit{In what year did the republican revolution take place?}
    
\item Correctness: The answer has to be correct.
    
    \badmark\ \textit{When was slavery abolished? 1761}\footnote{1761: Only trafficking was abolished.}\\    
    \goodmark\ \textit{When was slavery abolished? 1869}
    
\item Context independence: The correctness of the answer cannot depend on additional context that is not present in the question.
    
    \badmark\ \textit{What is the trend in the interior of the country?}\\    
    \goodmark\ \textit{Which soccer player has scored the most goals for the national team?}
    
\item Not a yes/no question: The question cannot be answerable with a simple yes/no.
    
    \badmark\ \textit{Is sardine a species of fish caught off the coast of our country?}
    
\end{enumerate}

\subsection{Endogenous point of view}
\label{Sect:pointOfview}

Another subset of guidelines include constraints related to the intended content and point of view. They seek to avoid the drawbacks discussed above, including stereotypes that emerge from points of view that are external to the culture at stake. They seek also to avoid getting focused only into so-called ``high culture", typically documented in encyclopedic knowledge sources. 

In the next section, as part of their empirical validation, we report on the application of the whole set of guidelines in the development of a dataset. 
To make it concrete and operational, we consider a given culture. In that use case, we consider the Portuguese culture in Portugal, and that is the reason why, in the guidelines below, there is a reference to it. 
Naturally, these guidelines are apt to support the development of similar datasets for any other particular culture (e.g.~related to countries, regions, professional groups, communities of practice, etc.) provided the respective reference to it replaces the reference to the Portuguese culture in these guidelines.

\begin{enumerate}[E1.]\small

\item Language: The question-answer pair should be in Portuguese, as it is used in Portugal.
 
    \badmark\ \textit{How long does it take to travel by bus [PT-BR: ônibus] between the capital and the second largest city?}\\    
    \goodmark\ \textit{How long does it take to travel by bus [PT-PT: autocarro] between the capital and the second largest city?}
    
\item Scope: The question-answer pair should be specific to the Portuguese culture.
    
    \badmark\ \textit{What is the number of electrons in nitrogen?}\\    
    \badmark\ \textit{Where are the biggest waves in the world surfed? Nazaré, Portugal.}\\    
    \goodmark\ \textit{How many World Cups has the national soccer team won?}
    
\item Knowledge level: The answer should be known by anyone that has grown up and get educated in Portugal.
    
    \badmark\ \textit{Who was the founder of Diário de Notícias?}\\    
    \goodmark\ \textit{Who was the first king?}
    
\item Domains: Avoid stereotypes, slang, etc.
    
    \badmark\ \textit{What do people usually say about politics?}
    
\item Temporal horizon: The question should be answerable on the basis of information that is at least three years old.
    
    \badmark\ \textit{Who is the current prime minister?}\\    
    \goodmark\ \textit{Who was the prime minister during the COVID-19 pandemic?}
    
\end{enumerate}

As can be inferred from these guidelines, in order to elicit the intended endogenous point of view, no indication is provided of what should be considered culture, or even Portuguese culture, leaving it up to the annotators to provide an operational substantiation of it with their contributed entries.

\subsection{Enhanced discriminative power}
\label{Sect:power}

The last subset of guidelines serves to avoid the limitations discussed above related to the insufficient discriminatory power of the resulting datasets, which runs counter to the ultimate goal that these datasets are intended to help achieve.

\begin{enumerate}[D1.]\small

\item Pragmatic context: The question should be posed naturally, as in a casual conversation, without having to specify that it pertains to Portugal or Portuguese culture, and avoiding proper names.
    
    \badmark\ \textit{Which Portuguese city is known as the “Venice of Portugal”?}\\    
    \goodmark\ In what year did the republican revolution take place?
    
\item Bigtech chatbots failure: Ideally, ChatGPT and similar models should give the wrong answer to the question.
    
    \badmark\ \textit{Who is the athlete known as the “black panther”?}\\    
    \goodmark\ \textit{In the 20th century, in which year did the first legislative elections take place after the dictatorship?}
    
\item Discriminative among cultures of Portuguese-speaking countries: The question should have an answer that is different from the answer that would be given by Portuguese speakers from other Portuguese-speaking countries, with other national cultures.
    
    \badmark\ \textit{What was the name of the dictatorship that ruled for most of the 20th century?}\\    
    \goodmark\ \textit{When was the most recent democratic regime established?}
    
\end{enumerate}

\section{Dataset for cultural alignment}
\label{Sect:dataset}

Following the guidelines described in the preceding section, and in order to assess them, we developed a benchmark named \benchmark. 
This dataset consists of 327 question-answer entries in Portuguese and is aimed at assessing cultural alignment with the Portuguese culture.%
\footnote{The creation of the Tuguesice-PT benchmark required a great amount of manual work by various people. To protect its status as test set, and following similar practice adopted by other benchmarks, it is not freely available online where it could easily be automatically scrapped as training data for models. The benchmark is available upon justified request.} 

A total of 9~annotators, undergraduate students aged 19--25, working independently of each other, were hired to produce candidate question-answer pairs under the guidelines presented above.
The proposed candidate entries were then verified by a separate team of adjudicators, that removed eventual duplicates and double checked for the compliance with the guidelines.

Before the annotation proper started, an experimental round was performed to familiarize the annotators with the guidelines, after which their proposed entries were discussed with them with respect to their conformity to the guidelines.
This also allowed for the refinement of a few guidelines, by making their phrasing clearer and by improving the contrasting examples that accompany them. 
At regular intervals, during the annotation period, meetings were held with the team of annotators to keep the annotation procedure aligned with its aim.

\section{Empirical evaluation}
\label{Sect:assessment}

\begin{table*}[!tp]
    \centering
    \begin{tabular}{l rrr rrr}
        \toprule
                         & \multicolumn{3}{c}{\benchmark} 
                         & \multicolumn{3}{c}{BLEnD-PT} \\
        model            & plain & oracle & $\Delta$
                         & plain & oracle & $\Delta$ \\
        \cmidrule(r){1-1}\cmidrule(lr){2-4}\cmidrule(l){5-7}
        gemini-2.5-flash & 35.78 & 77.68 & 42 &   55.17 & 54.74 &  0 \\
        gervasio70b      & 39.76 & 60.86 & 21 &   51.72 & 53.45 &  2 \\
        llama70b         & 25.69 & 63.30 & 38 &   50.00 & 51.29 &  1 \\
        mistral24b       & 15.60 & 57.19 & 42 &   43.97 & 46.98 &  3 \\
        gervasio8b       & 11.31 & 38.53 & 27 &   41.81 & 42.67 &  1 \\
        llama8b          &  9.79 & 40.37 & 31 &   40.52 & 43.97 &  3 \\
        sabia7b          &  7.95 & 27.83 & 20 &   19.83 & 31.03 & 11 \\
        \bottomrule
    \end{tabular}
    \caption{Results (accuracy, as a percentage), for \benchmark\ and for BLEnD-PT. Prompt formed by the question only (plain), and prompt formed by question preceded by the scope instruction (oracle). Difference between plain and oracle scores shown as $\Delta$.}
    \label{tab:results}
\end{table*}

Focusing on the development of datasets for cultural alignment, we conducted a series of experiments—--reported in this section—--to empirically assess both our analysis of the mainstream approach and its identified limitations, as well as the alternative approach we propose.

To enable a contrastive study, we put side by side the dataset described above---developed according to our proposed guidelines---with a second dataset for the same language and culture, constructed by following a mainstream design. We created this second dataset by applying the procedures used in BLEnD for other languages.

Accordingly, after machine translating a portion of the BLEnD questions into Portuguese, the outcome was manually revised, and the respective correct answers were associated to them, by the same team of highly skilled adjudicators of our dataset \benchmark. 
The resulting BLEnD-PT comprises 232 question-answer pairs.%
\footnote{Similarly to Tuguesice-PT, BLEnD-PT is available upon justified request.}

For the empirical study, we experimented with a range of models featuring diverse characteristics. 
To enhance a first crucial contrastive dimension, we used models that have been specifically fine-tuned for Portuguese side by side with models that were not and that are comparable to the former \textit{ceteris paribus}. We resorted to the Gervásio models, which have been fine-tuned for Portuguese \citeplanguageresource{santos-etal-2024-advancing}, and the generic Llama models that served as the starting points for their fine-tuning \citeplanguageresource{touvron2023llamaopenefficientfoundation}.

In order to assess also the possible effect of model size, we resorted to models over a range of sizes. 
The Gervásio and Llama models mentioned were used in their version 8B and 70B (Billion of parameters). 
To these we added the open model Mistral, with 24B parameters \citeplanguageresource{Mistral24B}, and the commercial, closed model Gemini 2.5 Flash \citeplanguageresource{2025gemini20to25flash}, a mixture-of-experts whose size is not disclosed by Google.

To help assess the eventual capacity to discriminate between quite close variants of the same language, besides the Gervásio, for European Portuguese, we also used the Sabiá 7B model, fine-tuned for Brazilian Portuguese \citeplanguageresource{sabia2023}.

As expected answers are short (given the questions included in the datasets), output generated by the models was limited to 64 tokens. 
The open models were run quantized at 4-bit for the sake of efficiency.
Gemini was run with thinking turned off.

As discussed in our analysis above, in the mainstream approach, questions tend to contain ``scope information'' (e.g.~name of the culture, country, proper names associated to a culture, etc.) that bias models into a specific culture and helps inflate their scores with respect to their alignment to that culture. 
To assess the possible effect of such ``oracle'', external to the questions themselves, we repeat the testing of the models by providing them with a system prompt, that precedes each question in run time, stating that the model should assume the context of Portuguese culture.\footnote{The prompt is presented in Appendix~\ref{appendix:prompt}.}

To enable the automatic evaluation of results under the accuracy metric, correctness of each answer by a model is established by checking whether the lower-cased gold answer string in the test dataset is contained in the lower-cased answer string output by the model.

Accuracy scores are compiled in Table~\ref{tab:results}.
The group of three columns on the left concern \benchmark\ and those on the right concern BLEnD-PT.
In each pair of columns with accuracy scores, the column on the left shows scores obtained with a ``plain'' prompt, formed by the question only, and the column in the right the scores obtained by prompting the models with the ``external oracle''.
The difference between the scores in these column pairs is shown in the third column $\Delta$.

\subsection*{Discussion}

These evaluation results demonstrate that, in comparison to the mainstream approach, and as it was argued and sought for, our proposed approach for the design of datasets to assess cultural alignment ensures more discriminative power, along the different dimensions experimented with.

The ``plain'' scores under \benchmark\ are more distant among themselves for the models fine-tuned into Portuguese and their non-finetuned base versions:
The gaps from Gervásios 70B and 8B to Llama 70B and 8B are 14.07 and 1.52 p.p., respectively, while these gaps are only 1.72 and 1.29, respectively, under the competitor dataset. 
This indicates that our \benchmark\, and thus our proposed guidelines, are better at discriminating between models specialized and non-specialized in a given culture. 

Also, the ``plain'' scores under our approach are in general lower. 
The top value 39.76, for Gervário 70B, is lower than the top 55.17, for Gemini, under the representative dataset of mainstream approach.
This indicates that \benchmark\ is less exhausted, and is thus more challenging and usable to assess cultural alignment for a longer period of experimentation, with models with ever increasing performance levels.

The evaluation results also demonstrate the magnitude of the undesirable biasing effect by an ``oracle'' (conveyed in the system prompt) that is external to the inner capabilities of the model. 
The scores end up inflated by a value from 20 p.p., with Sabiá 7B, to as much as 42~p.p., with Gemini, as it can be observed in the first $\Delta$-column, under \benchmark. 
The scores in the second $\Delta$-column, in turn, show how much this inflating effect is almost completely obfuscated there, with deltas ranging between 0~p.p., for Gemini, and 3~p.p., for Llama 8B.
The contrast between the two columns of deltas demonstrate in a striking fashion how mainstream approaches datasets may not be able to make evident the lack of alignment of models to a given culture. 
In this respect, it is very eloquent the contrast between the delta of 42~p.p. for Gemini under \benchmark\ and its delta of 0~p.p. under the other approach---suggesting the possible alignment of this model with this culture is at its peak, which the larger delta made evident by our approach fully debunks.\footnote
{
Naturally, these results also confirm that, in general, larger models perform better than smaller ones. 
That is as expected---given larger models are trained on larger amounts of data---and it is orthogonal to the argument of the present study.
}

\section{Conclusions}
\label{Sect:conclusions}

We presented a review of existing approaches to the design and development of datasets for assessing the alignment of LLMs with respect to a given culture and identified limitations for them. 

To address these issues, we proposed a set of design guidelines for annotators, and reported on the construction of a dataset that we developed in accordance to these principles---though for the sake of the feasibility of our study, the dataset covers one language, test datasets can be developed along these guidelines for any language. 

With this in place, we reported on a series of experiments we conducted with this dataset and a range of LLMs. 
The results obtained demonstrate that our proposed design informs the construction of test datasets with greater discriminative power, effectively distinguishing between models specialized for a given culture and those that are not, thus empirically validating its adoption for the construction of further datasets for further languages.

\section*{Acknowledgments}

This research was partially supported by:
ACCELERAT.AI---Multilingual Intelligent Contact Centers, funded by PRR---Plano de Recuperação e Resiliência, from Portugal, through IAPMEI (C625734525-00462629);
PORTULAN CLARIN---Research Infrastructure for the Science and Technology of Language, funded by LISBOA2030 (FEDER-01316900); 
Hey, HAL, curb your hallucination!, funded by FCT---Fundação para a Ciência e Tecnologia (2024.07592.IACDC); 
and LLMs4EU---Large Language Models for the European Union, funded by the DIGITAL Programme (DIGITAL-2024-AI-06-LANGUAGE-01).

\section{Bibliographical References}
\bibliographystyle{lrec2026-natbib}
\bibliography{refs-bibliography}

\section{Language Resource References}
\bibliographystylelanguageresource{lrec2026-natbib}
\bibliographylanguageresource{refs-languageresources}

\appendix
\section{The ``oracle'' prompt}
\label{appendix:prompt}

The ``oracle'' prompt, which states that the model should assume the context of Portuguese culture, is as follows:
\begin{quote}
  Assume que és uma pessoa portuguesa, nascida e criada em Portugal. Assume que a pergunta se refere a Portugal e à sua cultura. Falas a língua portuguesa tal como é usada em Portugal. A tua gramática e o teu vocabulário é o da língua portuguesa como ela é falada em Portugal. A tua moeda é o euro.  
\end{quote}

which translates to 
\begin{quote}
Assume you are a Portuguese person, born and raised in Portugal. Assume that the question refers to Portugal and its culture. You make use of the Portuguese language as it is used in Portugal. Your grammar and vocabulary are those of the Portuguese language as it is used in Portugal. Your currency is the Euro.    
\end{quote}

\end{document}